\documentclass[10pt,twocolumn,letterpaper]{article}

\usepackage{multirow} 
\usepackage{color}
\usepackage{pifont}

\usepackage{siunitx,calc,xcolor}

\usepackage[table]{xcolor}

\usepackage[accsupp]{axessibility}  
\newlength{\basexsep}
\newlength{\basexwidth}
\usepackage{amsmath}
\usepackage{bm}
\newcommand{\basex}[1]{\textcolor{gray!45}{\footnotesize \ $\uparrow${#1}}}
\newcommand{\basexx}[1]{\textcolor{gray!85}{\footnotesize \ $\uparrow${#1}}}

\newcommand{\up}[1]{\textcolor{OliveGreen}{\footnotesize \ $\uparrow$\textbf{#1}}}

\newcommand{\mypara}[1]{\smallskip\noindent\textbf{#1}}

\usepackage{nicefrac}

\usepackage{cvpr}              
\definecolor{cvprblue}{rgb}{0.21,0.49,0.74}
\usepackage[pagebackref,breaklinks,colorlinks,allcolors=cvprblue]{hyperref}

\def\paperID{15711} 
\def\confName{CVPR}
\def\confYear{2026}

\title{Beyond Binary Contrast: Modeling Continuous Skeleton Action Spaces \\ with Transitional Anchors}

\author{
    Yingjie Feng$^{1}$\footnotemark[1] \quad 
    Yi Wang$^{2}$\footnotemark[1] \quad
    Jiaze Wang$^3$ \quad
    Anfeng Liu$^2$ \quad
    Zhuotao Tian$^{1}$\footnotemark[2]
    \vspace{10pt}
    \\
    $^1$ Harbin Institute of Technology, Shenzhen \quad
    $^2$ Central South University
    \\
    $^3$ FitX Technology (Hong Kong) Limited
}

\begin{document}
\maketitle
\footnotetext[1]{Equal contribution.} 
\footnotetext[2]{Corresponding author (tianzhuotao@hit.edu.cn).}
\begin{abstract}
Self-supervised contrastive learning has emerged as a powerful paradigm for skeleton-based action recognition by enforcing consistency in the embedding space. However, existing methods rely on binary contrastive objectives that overlook the intrinsic continuity of human motion, resulting in fragmented feature clusters and rigid class boundaries. To address these limitations, we propose \textbf{TranCLR}, a Transitional anchor-based Contrastive Learning framework that captures the continuous geometry of the action space. Specifically, the proposed Action Transitional Anchor Construction (ATAC) explicitly models the geometric structure of transitional states to enhance the model's perception of motion continuity. Building upon these anchors, a Multi-Level Geometric Manifold Calibration (MGMC) mechanism is introduced to adaptively calibrate the action manifold across multiple levels of continuity, yielding a smoother and more discriminative representation space. Extensive experiments on the NTU~RGB+D, NTU~RGB+D~120 and PKU-MMD datasets demonstrate that TranCLR achieves superior accuracy and calibration performance, effectively learning continuous and uncertainty-aware skeleton representations. The code is available at \url{https://github.com/Philchieh/TranCLR}.
\end{abstract}    
\vspace{-1mm}
\section{Introduction}
Skeleton data provides a concise yet informative representation of human joints, naturally reflecting motion dynamics. Skeleton-based action recognition has become a key research topic in computer vision~\cite{zhu2025semantic, wang2024skeleton, chen2024neuron, wang2025foundation, qu2024llms, liu2024recovering}, benefiting from recent advances in deep learning and large-scale 3D representation learning~\cite{wu2024ppt, zhang2025concerto, peng2024oa, wang2024groupcontrast, ning2023boosting}. Moreover, dense perception and open-vocabulary vision-language models enable richer understanding of human motion and multimodal context~\cite{tian2022adaptive, tian2023learning, shao2024explore, wang2025declip, wang2025generalized, lai2024lisa, li2025perception, yang2024unified, tian2022generalized}, with applications in human-computer interaction, medical rehabilitation, and video understanding.

\begin{figure}[t]
    \centering
        \includegraphics[width=\linewidth]{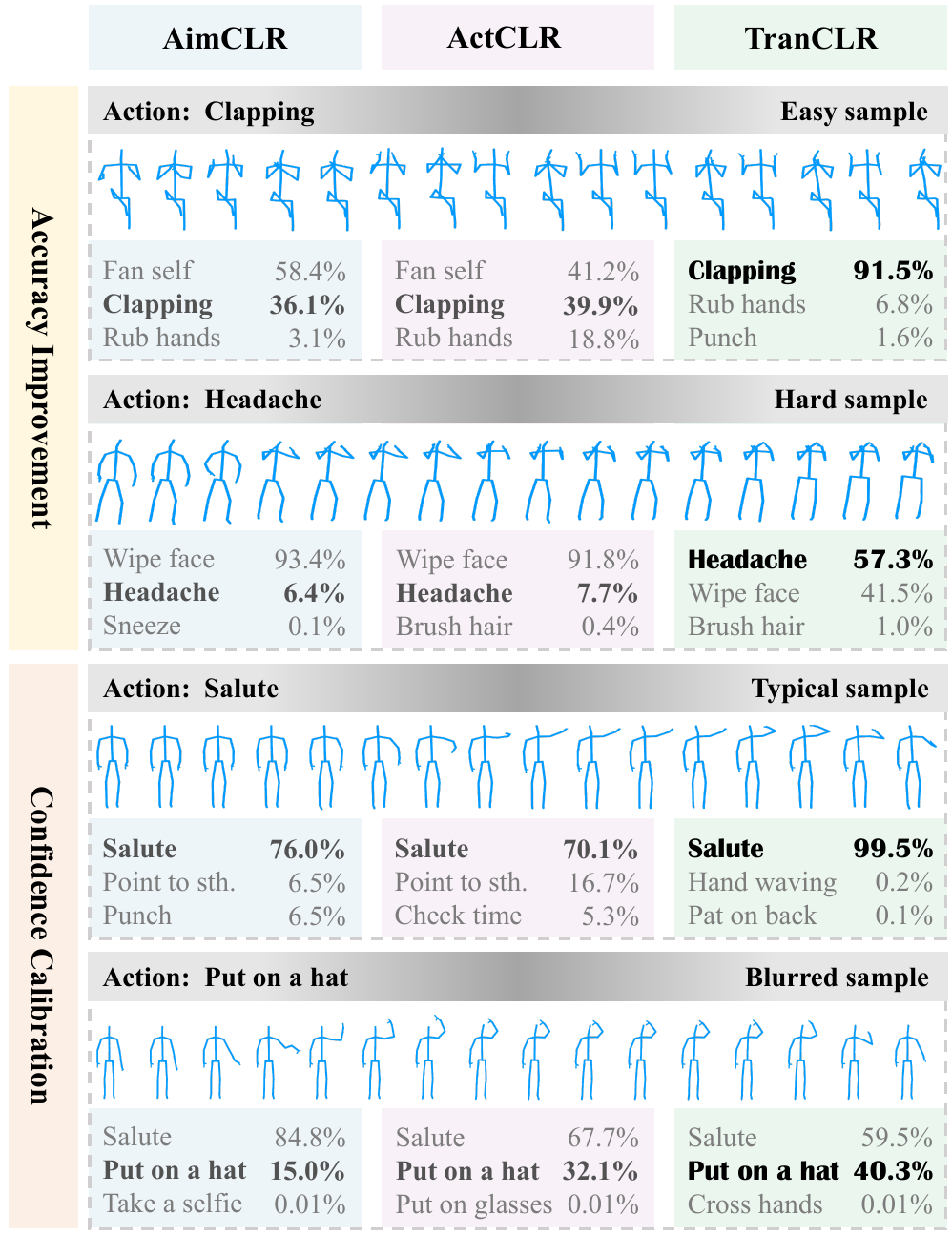}
\caption{\textbf{Illustration of TranCLR’s advantages.} Our method achieves higher accuracy on challenging samples and better-calibrated confidence across diverse motion scenarios. The top-3 predictions with confidence scores are shown for each sample.} 
    \label{fig: intro}
    \vspace{-14pt}
\end{figure}

Recent self-supervised methods~\cite{abdelfattah2024maskclr, zhang2024self, chen2022contrastive} have emerged as the dominant paradigm for skeleton representation, alleviating the reliance on large-scale annotations. Among them, contrastive learning~\cite{lin2023actionlet, lin2025self, guo2022contrastive, zhang2023hierarchical} achieves remarkable success by constructing positive and negative pairs to enforce embedding consistency. However, most methods adopt a binary objective: positives are pulled together while negatives are pushed apart. While effective for separability, it may overlook the intrinsic continuity of human motion, where actions evolve gradually and often share sub-motions. This inherent characteristic of natural human action raises a critical question: \textbf{\emph{``Can binary contrast faithfully capture the continuity of human motion?''}}

To investigate this, we visualize top-3 predictions with confidence scores from various contrastive learning-based models (\eg, AimCLR~\cite{guo2022contrastive}, ActCLR~\cite{lin2023actionlet}) on multiple motion samples. As shown in \cref{fig: intro}, such methods often produce unreliable and uncalibrated predictions, especially for hard and blurred action samples. These observations suggest that binary contrastive objectives provide only coarse similarity signals, lacking fine-grained distance perception in the embedding space. The limitation is reflected in two key aspects: \textbf{(1) Limited intra-class connectivity.} Positive pairs are typically derived from simple augmentations of the same sample, leaving distinct samples of the same action unpaired and fragmenting intra-class clusters. \textbf{(2) Rigid inter-class boundaries.} Actions with shared sub-motions are strictly separated by binary negatives, disrupting the smooth topology of the motion manifold. This rigid separation poses challenges for modeling transitional or ambiguous behaviors, which in turn affects the model’s calibration and confidence estimation.

To overcome these limitations, we advocate for a transition from discrete binary contrast to a more continuous and topology-aware representation paradigm. Instead of merely distinguishing between similar and dissimilar instances, we aim to capture the latent continuity that inherently exists between human actions. In this paper, we introduce \textbf{TranCLR}, a novel Transitional anchor-based Contrastive Learning framework, which constructs smoother and more topology-aware action spaces for universal human motion understanding. We first introduce an Action Transitional Anchor Construction (ATAC) strategy to explicitly characterize intermediate motion states. Rather than representing physically interpretable actions, transitional anchors serve as manifold regularizers, enriching the latent topology with intermediate states that lie along plausible semantic paths between actions. These anchors capture both semantic smoothness and local spatial-temporal fidelity, enabling the model to better discriminate fine-grained actions and reason over ambiguous or transitional behaviors.

\begin{figure}[t]
    \centering
        \includegraphics[width=\linewidth]{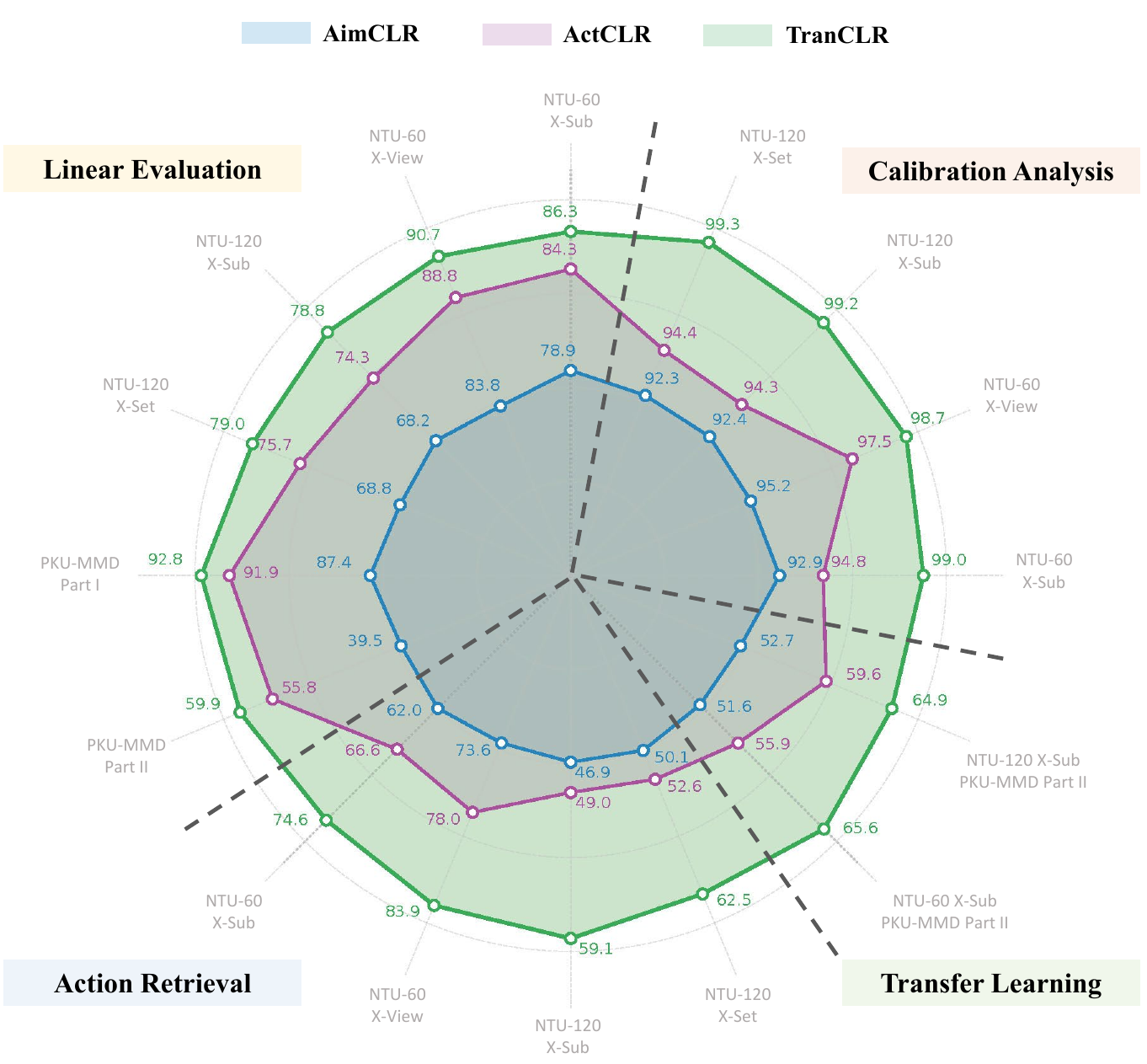}
\caption{\textbf{Performance gains of TranCLR across various tasks.} TranCLR surpasses AimCLR and ActCLR in Linear Evaluation, Transfer Learning, Skeleton-Based Action Retrieval, and Calibration Analysis, demonstrating superior overall performance; calibration error metrics are reversed for consistent direction.} 
    \label{fig: performance}
    \vspace{-14pt}
\end{figure}

Building upon these transitional anchors, we further introduce a Multi-Level Geometric Manifold Calibration (MGMC) mechanism that promotes smooth feature evolution across varying levels of motion continuity. By aligning transitional anchors with their mixed embeddings, this mechanism mitigates the rigid separability caused by binary comparisons. Moreover, it regulates the relative distances among transitional anchors to refine the manifold geometry. Through this self-regularization process, MGMC preserves global topological coherence, resulting in a smooth and discriminative representation space across scales.

Extensive experiments demonstrate that TranCLR consistently outperforms state-of-the-art methods across various settings, illustrated in \cref{fig: performance}. Specifically, TranCLR improves the average accuracy of the baseline by 17.2\% on NTU~RGB+D~120 dataset~\cite{shahroudy2016ntu, liu2020ntu}. In transfer learning, TranCLR achieves 65.6\% accuracy on the challenging PKU-MMD Part II dataset~\cite{liu2020a}. Furthermore, results on action retrieval and calibration analysis indicate that TranCLR not only exhibits strong robustness and generalization capability but also provides reliable uncertainty estimates, enhancing action understanding in open-world scenarios.

Our main contributions can be summarized as follows:

\begin{itemize}[leftmargin=0.7cm]
\item We propose TranCLR, a novel anchor-based contrastive learning framework that shifts from discrete binary similarity modeling to a continuous and topology-aware representation paradigm for skeleton-based action understanding.

\item We design a multi-level geometric manifold calibration mechanism that progressively aligns and regulates feature distances among anchors, promoting smoothness and discriminability of the representation space.

\item TranCLR not only exhibits strong robustness and generalization, but also provides reliable uncertainty estimation for ambiguous actions, achieving superior performance across linear evaluation, transfer learning, action retrieval, and calibration analysis.
\end{itemize}

\section{Related Work}

\mypara{Skeleton-Based Action Recognition.} Skeleton-based action recognition classifies human actions using skeletal joint trajectories. Early deep-learning approaches used RNNs to model temporal dynamics \cite{du2015hierarchical, liu2017global} or CNNs by treating skeleton sequences as pseudo-images \cite{ke2017anew, liu2017enhanced}. However, RNN-based methods struggle with optimization and long-range dependencies, while CNN-based ones overlook the skeleton's inherent graph topology. The introduction of the Spatial-Temporal Graph Convolutional Network (ST-GCN) by Yan \etal~\cite{yan2018spatial} marked a pivotal moment in the field. By representing the skeleton as a graph and applying graph convolutions, this approach was the first to explicitly model the non-Euclidean connectivity of body joints. This graph-based paradigm has since become the \textit{de facto} standard, inspiring numerous subsequent works that refine the spatio-temporal modeling capabilities \cite{shi2019twostream, cheng2020shiftgcn, chen2021channel, zhou2024blockgcn, liu2025revealing}. More recently, Transformer-based architectures \cite{plizzari2021skeleton, shi2020decoupled} have demonstrated strong performance in global context modeling, yet their substantial computational and data requirements often render them impractical. Given the proven efficacy and efficiency of graph-based models, our work is built upon the foundational ST-GCN architecture.

\mypara{Skeleton-Based Contrastive Learning.} While supervised skeleton-based action recognition achieves high accuracy, it depends on large-scale annotated datasets. Self-supervised learning alleviates this by deriving robust representations from unlabeled data, with contrastive learning as the prevailing paradigm~\cite{chen2020simple, he2020momentum, grill2020bootstrap, chen2021exploring}. This method constructs an embedding space by attracting positive pairs and repelling negative ones. Early works, such as SkeletonCLR~\cite{li20213d}, adapted instance discrimination frameworks like MoCo v2~\cite{chen2020mocov2} for skeleton data. Subsequent works exploit skeletal modality richness: CrosSCLR~\cite{li20213d} leverages cross-view learning, while CMD~\cite{Mao_2022_CMD} employs bidirectional mutual distillation for representation refinement. Another key focus is generating harder positive samples via extreme augmentations~\cite{guo2022contrastive}, hierarchical schedules~\cite{zhang2023hierarchical}, or spatio-temporal mixing of sequences~\cite{chen2022contrastive}. Further efforts prioritize semantic information. Hu \etal~\cite{hu2024global} proposed a global-to-local contrastive strategy, and ActCLR~\cite{lin2023actionlet} explicitly segmented skeletons into discriminative \emph{actionlets} to enforce semantic invariance. More recently, hybrid paradigms have begun to combine masked modeling with contrastive objectives: STARS~\cite{mehraban2025stars} couples masked prediction with nearest-neighbor contrastive tuning, while NAT-CL~\cite{tao2025rethinking} shows that masked reconstruction pretraining can be further strengthened by contrastive learning. However, existing methods rely on binary contrastive objectives that overlook the continuity of human motion, yielding rigid decision boundaries and weaker modeling of transitional behaviors.
\begin{figure*}[t]
    \centering
        \includegraphics[width=\linewidth]{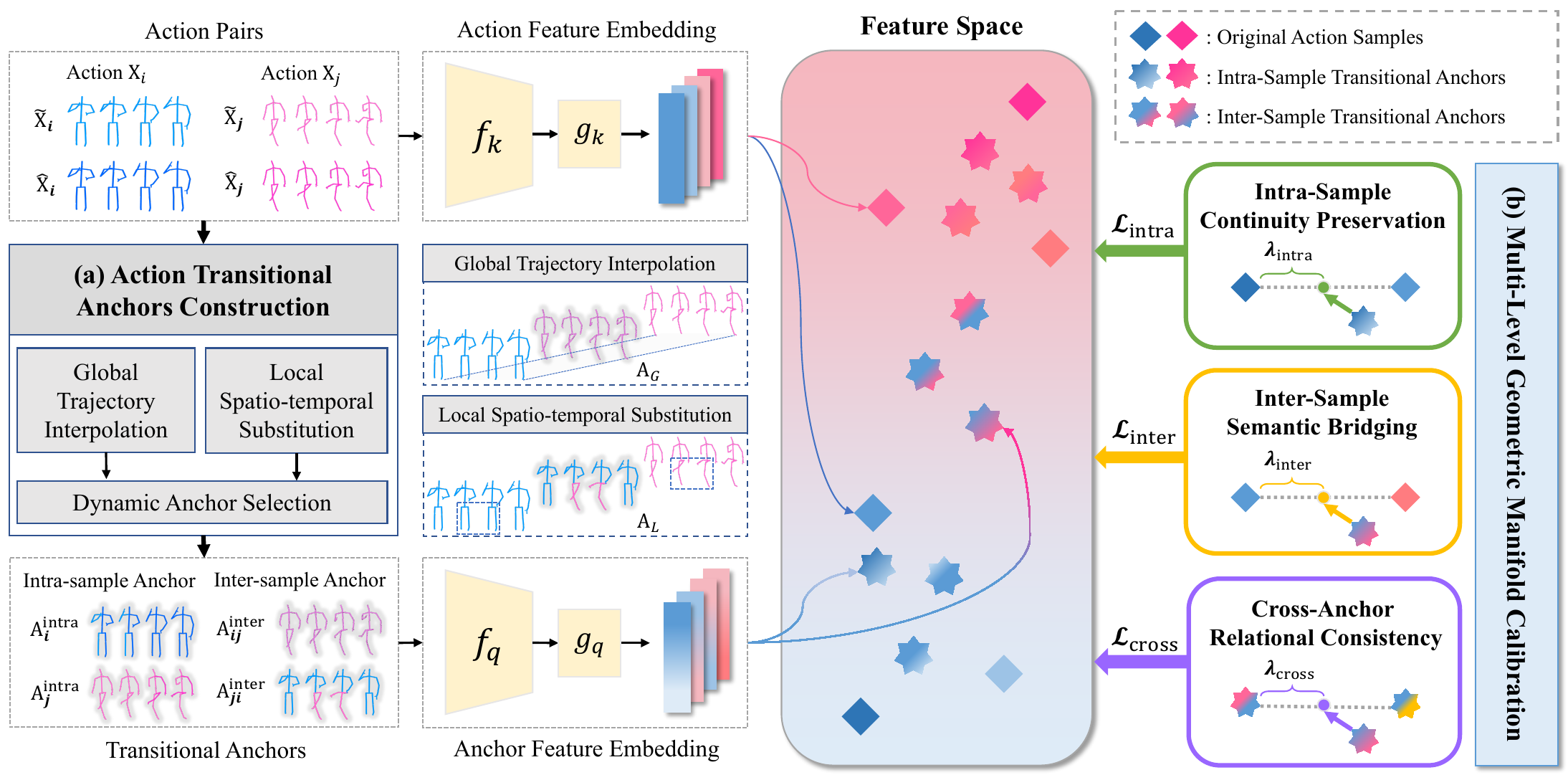}
\caption{\textbf{The overall scheme of TranCLR.} (a) Action transitional anchors are constructed through global trajectory interpolation and local spatio-temporal substitution to generate intermediate motion states that enrich the manifold topology. (b) Based on these anchors, the Multi-Level Geometric Manifold Calibration (MGMC) progressively aligns and regulates feature distances, producing a smooth, topology-consistent and robust skeleton representation space. } 
    \label{fig2}
    \vspace{-10pt}
\end{figure*}

\section{Methodology}
To address the above issues, we propose \emph{TranCLR}, a novel approach for a well-structured action representation manifold through geometric calibration. We first introduce the Action Transitional Anchor Construction (ATAC) strategy in \cref{subsec3.2}, which constructs transitional anchors. We then detail our Multi-Level Geometric Manifold Calibration (MGMC) mechanism in \cref{subsec3.3}, which leverages these anchors to enforce a coherent structure in the latent space. Finally, in \cref{subsec3.4}, we present a complementary soft alignment strategy and the unified training objective. The overall scheme is illustrated in \cref{fig2}.

\subsection{Preliminary}
Our approach is built upon MoCo v2~\cite{chen2020mocov2}. The framework maintains two networks: an \emph{online} network (encoder $f_q$ and projector $g_q$) updated by back-propagation, and a \emph{momentum} network (encoder $f_k$ and projector $g_k$) whose parameters follow an exponential moving average of the online network. For brevity, we denote the composed encoder-projector as $h(\cdot)=g(f(\cdot))$. Given two augmented views of an input sample $\widehat{\mathbf{X}}_i$ and $\widetilde{\mathbf{X}}_i$, the online network produces a query representation $\widehat{\boldsymbol{z}}_i = h_q(\widehat{\mathbf{X}}_i)$, while the momentum network generates a key representation $\widetilde{\boldsymbol{z}}_i = h_k(\widetilde{\mathbf{X}}_i)$. The model is trained using the InfoNCE~\cite{oord2018representation} loss:
\small
\begin{equation*} 
\mathcal{L}= -
\log \frac{\exp\!\left(\mathrm{sim}(\widehat{\boldsymbol z}_i, \widetilde{\boldsymbol z}_i)/\tau\right)}
{\displaystyle \exp\!\left(\mathrm{sim}(\widehat{\boldsymbol z}_i, \widetilde{\boldsymbol z}_i)/\tau\right) \!+ \!\sum_{\boldsymbol m_j\in \mathcal M} \!\exp\!\left(\mathrm{sim}(\widehat{\boldsymbol z}_i, \boldsymbol m_j)/\tau\right)},
\end{equation*} 
\normalsize
where $\mathrm{sim}(\cdot,\cdot)$ is the cosine similarity, $\tau$ is a temperature hyperparameter and $\mathcal{M}$ is a dynamic queue of negative samples maintained by the momentum network. We adopt this architecture as the foundational basis for our method.


\subsection{Action Transitional Anchor Construction}
\label{subsec3.2}

To create a more structured action representation space that captures continuous transitions rather than simple binary comparisons, we propose the Action Transitional Anchor Construction (ATAC) strategy. ATAC models the geometric manifold of action transitions by synthesizing \emph{Transitional Anchors}, intermediate waypoints between action samples $\mathbf{X}_i$ and $\mathbf{X}_j$ that encode their relationships and serve as manifold regularizers rather than ground-truth poses. This encourages a topologically coherent action manifold reflecting the continuous nature of human actions.

\mypara{Global Trajectory Interpolation.}
Inspired by Mixup~\cite{zhang2017mixup}, we model transitions via linear interpolation to create a globally smooth trajectory on the data manifold. The global anchor $\mathbf{A}_{G}$ is a convex combination:
\begin{equation*} 
\mathbf{A}_{G} = \lambda_G \mathbf{X}_i + (1-\lambda_G) \mathbf{X}_j,
\end{equation*} 
where $\lambda_G \sim U(0, 1)$ serves as an explicit measure of its semantic distance to $\mathbf{X}_i$ and $\mathbf{X}_j$.

\mypara{Local Spatio-temporal Substitution.} 
Global interpolation may blur fine-grained motion details. To preserve kinematics while ensuring plausibility, we introduce a local spatio-temporal substitution method at the body-part level, inspired by \cite{zhang2024shapmix, liu2023skeletoncutmix}. We decompose the skeleton into five parts: $\mathcal{P}=\{$\emph{left-arm}, \emph{right-arm}, \emph{left-leg}, \emph{right-leg}, \emph{trunk}$\}$. 

Specifically, we randomly select $\mathcal{S} \in [\mathcal{S}_{\min}, \mathcal{S}_{\max}]$ body parts and a time window of length $\mathcal{T} \in [\mathcal{T}_{\min}, \mathcal{T}_{\max}]$. Using $\mathbf{X}_j$ as the base, we extract a sub-sequence of duration $\mathcal{T}' \in [\kappa_l \mathcal{T}, \kappa_r \mathcal{T}]$ from the corresponding parts in $\mathbf{X}_i$, resize it to $\mathcal{T}$ frames, and substitute it into the selected regions of $\mathbf{X}_j$. The resulting local anchor $\mathbf{A}_L$ is formulated as:
\begin{equation*}
    \mathbf{A}_{L} = \mathbf{M} \odot \text{Resize}(\mathbf{X}_i \odot \mathbf{M}', \mathcal{T}) + (\mathbf{1} - \mathbf{M}) \odot \mathbf{X}_j,
\end{equation*} 
where $\mathbf{M} \in \{0,1\}^{|\mathcal{P}| \times T}$ masks substitution regions, and $\mathbf{M}'$ isolates the extraction from $\mathbf{X}_i$. The mask mean $\lambda_L = \mathbb{E}[\mathbf{M}]$ implicitly quantifies the semantic distance. 

\mypara{Dynamic Anchor Selection.} 
To harness the complementary benefits of semantic smoothness and spatial-temporal fidelity, we dynamically select the transitional anchor generated by global interpolation and local substitution with equal probability for each action pair $(\mathbf{X}_i, \mathbf{X}_j)$:
\begin{equation*}
\scalebox{0.9}{$
\text{ATAC}(\mathbf{X}_i, \mathbf{X}_j; \lambda) = \mathbf{1}_{\{p<0.5\}}\,\mathbf{A}_{\mathrm{G}}
+ \mathbf{1}_{\{p\ge0.5\}}\,\mathbf{A}_{\mathrm{L}}, \quad
p \sim U(0,1)
$}
\end{equation*} 
where $\lambda$ corresponds to either $\lambda_G$ or $\lambda_L$, representing semantic distance. This exposes the model to diverse transitions for a continuous, fine-grained representation space.


\subsection{Multi-Level Geometric Manifold Calibration}
\label{subsec3.3}

To endow the latent space with a coherent geometric structure, we introduce the Multi-Level Geometric Manifold Calibration (MGMC) mechanism. This calibration strategy leverages transitional anchors from ATAC to enforce topological consistency at three complementary levels, transforming the representation space from a simple collection of points into a smooth, semantically ordered manifold. Building on a contrastive learning paradigm, we generate two augmented views per sample in each mini-batch: $\{\widehat{\mathbf{X}}_i\}_{i=1}^N$ and $\{\widetilde{\mathbf{X}}_i\}_{i=1}^N$, forming the basis of our calibration.

\mypara{Intra-Sample Continuity Preservation.}
Standard contrastive methods treat augmented views of the same action as equivalent positives, potentially fragmenting intra-class continuity and creating disjoint clusters. To mitigate this, we enforce a homomorphic mapping: semantic transitions in the input space must correspond to linear interpolations in the embedding space, promoting local cohesion. 

Concretely, we construct an intra-sample transitional anchor $\mathbf{A}^{\text{intra}}_i = \text{ATAC}(\widehat{\mathbf{X}}_i, \widetilde{\mathbf{X}}_i; \lambda_{\text{intra}})$ as an intermediate state and align via the loss $\mathcal{L}_{\text{intra}}$:
\begin{equation*} 
h_q(\mathbf{A}^{\text{intra}}_i) \longleftrightarrow \lambda_{\text{intra}} h_k(\widehat{\mathbf{X}}_i) + (1-\lambda_{\text{intra}}) h_k(\widetilde{\mathbf{X}}_i).
\end{equation*} 
This ensures smooth interpolation along positive trajectories, mitigating discontinuities and providing a stable base for further calibrations.

\mypara{Inter-Sample Semantic Bridging.}
While intra-sample calibration ensures local continuity, the semantic boundaries between different action classes may remain rigid, impeding transitions between related yet distinct actions. To address this, we extend our framework to the inter-sample level, assuming actions lie on a manifold where distances reflect semantic similarity.

Specifically, for distinct actions $\mathbf{X}_i$ and $\mathbf{X}_j$, we create an inter-sample anchor $\mathbf{A}^{\text{inter}}_{ij} = \text{ATAC}(\widehat{\mathbf{X}}_i, \widehat{\mathbf{X}}_j; \lambda_{\text{inter}})$ as their semantic midpoint and align via the loss $\mathcal{L}_{\text{inter}}$:
\begin{equation*} 
h_q(\mathbf{A}^{\text{inter}}_{ij}) \longleftrightarrow \lambda_{\text{inter}} h_k(\widetilde{\mathbf{X}}_i) + (1-\lambda_{\text{inter}}) h_k(\widetilde{\mathbf{X}}_j).
\end{equation*} 
This operation enriches the manifold by representing gradual semantic transitions (\eg, \emph{walking} to \emph{running}) as continuous distances, which softens rigid class boundaries.

\mypara{Cross-Anchor Relational Consistency.}
While the first two calibration levels establish local continuity and semantic bridging, they generate numerous anchors from related parent pairs, leading to partial semantic overlap that remains geometrically unconstrained. These overlaps encode implicit topological relationships of the manifold’s global organization, which can be leveraged as a weak supervisory signal to refine structural coherence. Accordingly, our third level implements this by imposing geometric constraints on the anchors within these overlaps.

For a mini-batch of size $N$, we require a diverse set of anchor pairs that share parents. As a deterministic and efficient sampling strategy, we pair each sample $\mathbf{X}_i$ with its reverse-ordered counterpart $\mathbf{X}_{N-i+1}$ and generate two anchors using distinct mixing coefficients $\lambda_1$ and $\lambda_2$:
\begin{equation*}
\begin{aligned}
\mathbf{A}^{(1)}_{i} &= \text{ATAC}(\widehat{\mathbf{X}}_i, \widehat{\mathbf{X}}_{N-i+1}; \lambda_1), \\
\mathbf{A}^{(2)}_{i} &= \text{ATAC}(\widetilde{\mathbf{X}}_i, \widetilde{\mathbf{X}}_{N-i+1}; \lambda_2).
\end{aligned}
\end{equation*} 
We can naturally observe that the anchor $\mathbf{A}^{(1)}_{i}$ is related to two other anchors: its homologous partner $\mathbf{A}^{(2)}_{i}$ (same parent order) and its cross-homologous partner $\mathbf{A}^{(2)}_{N-i+1}$ (reverse parent order). To quantify their relationships, we define compositional similarity scores that assess the anchors' overlap in parentage. For each shared parent sample, we sum the minimum of the mixing coefficients applied in the two anchors. These scores are formally defined as:
\begin{equation*}
\scalebox{0.8}{$
\begin{alignedat}{3}
&\text{homo :}\;& \  k_{h}\!\left(\mathbf{A}^{(1)}_{i},\mathbf{A}^{(2)}_{i}\right)
&= \min(\lambda_1,\lambda_2) + \min(1-\lambda_1,1-\lambda_2),\\
&\text{cross :}\;& \ k_{c}\!\left(\mathbf{A}^{(1)}_{i},\mathbf{A}^{(2)}_{N-i+1}\right)
&= \min(\lambda_1,1-\lambda_2) + \min(1-\lambda_1,\lambda_2).
\end{alignedat}
$}
\end{equation*} 
These scores serve as a metric for the relative distances between the anchor $\mathbf{A}^{(1)}_{i}$ and its partners within the compositional space. Consequently, we normalize them to derive a weighting coefficient $\lambda_{\text{cross}}=k_h / (k_h+k_c)$ that reflects this proximity, and align via the loss $\mathcal{L}_{\text{cross}}$:
\begin{equation*}
\scalebox{0.9}{$
    h_q(\mathbf{A}^{(1)}_{i}) \longleftrightarrow \lambda_{\text{cross}} h_k(\mathbf{A}^{(2)}_{i}) + (1-\lambda_{\text{cross}} ) h_k(\mathbf{A}^{(2)}_{N-i+1}). 
$}
\end{equation*} 
While $\mathcal{L}_{\text{intra}}$ and $\mathcal{L}_{\text{inter}}$ ensure local and inter-sample smoothness, $\mathcal{L}_{\text{cross}}$ guarantees that the relative positions of the anchors themselves are coherent, refining the entire web of transitional relationships into a topologically consistent global action manifold.

\subsection{Soft Alignment and Overall Objective}
\label{subsec3.4}

The alignment objectives in \cref{subsec3.3} may introduce conflicts. For instance, $\mathcal{L}_{\text{intra}}$ promotes intra-class compactness by pulling augmented views tightly together, whereas $\mathcal{L}_{\text{inter}}$ fosters inter-class continuity by easing boundaries between distinct actions. Such tensions can induce training instability when using rigid hard alignments like InfoNCE~\cite{oord2018representation}.

\mypara{Soft Alignment.} To alleviate this, we propose a soft alignment strategy drawing from knowledge distillation. For each alignment query-target pair $(\boldsymbol{q}, \boldsymbol{k})$, we identify the top-$K$ most similar samples to $\boldsymbol{k}$ from the memory queue $\mathcal{M}$, storing their indices in $\mathcal{N}_K$. This selection retains only high-confidence neighbors semantically aligned with the target, which excludes noisy samples to preserve alignment signal purity and focus on meaningful local relationships, enhancing the stability and efficiency of training. We then compute similarity vectors:
\begin{equation*}
    \boldsymbol{p}_k = \left[ \text{sim}(\boldsymbol{k}, \boldsymbol{m}_j) \right]_{j \in \mathcal{N}_K}, \quad \boldsymbol{p}_q = \left[ \text{sim}(\boldsymbol{q}, \boldsymbol{m}_j) \right]_{j \in \mathcal{N}_K},
\end{equation*} 
and align their softened distributions using KL divergence with asymmetric temperatures:
\begin{equation*}
    \mathcal{L}(\boldsymbol{q}, \boldsymbol{k}) 
    = 
    \text{KL}
    \left( 
    \text{softmax}(\boldsymbol{p}_k / \tau_k) 
    \parallel 
    \text{softmax}(\boldsymbol{p}_q / \tau_q) 
    \right),
\end{equation*} 
where $\tau_k < \tau_q$ sharpens the target distribution to emphasize peak similarities. This encourages the query to adopt the target's sharpened affinity profile, fostering smooth semantic neighborhoods and consistent local manifold geometry.

\mypara{Training Objectives.} Applying this soft alignment to our three calibration levels yields $\mathcal{L}_{\text{intra}}$, $\mathcal{L}_{\text{inter}}$, and $\mathcal{L}_{\text{cross}}$. The unified objective is the sum of these components:
\begin{equation*}
        \mathcal{L} = \mathcal{L}_{\text{intra}} +  \mathcal{L}_{\text{inter}} + \mathcal{L}_{\text{cross}}.
\end{equation*} 
The model is optimized by minimizing this objective.
\begin{table*}[tb]
\small
\centering
\caption{Comparison of linear evaluation results for contrastive learning approaches on the NTU dataset.}
\begin{tabular}{l|c|ccc|ccc}
    \toprule
    \multirow{2}{*}{\textbf{Method}} & \multirow{2}{*}{\textbf{Stream}} 
    & \multicolumn{3}{c|}{\textbf{NTU RGB+D}} 
    & \multicolumn{3}{c}{\textbf{NTU RGB+D 120}} \\
    & & X-Sub & X-View & Avg. & X-Sub & X-Set & Avg. \\
    \midrule

    SkeletonCLR~\cite{li20213d} & Joint & 68.3 \phantom{\basex{0.0\phantom{0}}} & 76.4 \phantom{\basex{0.0\phantom{0}}} & 72.4 \phantom{\basex{0.0\phantom{0}}} & 56.8 \phantom{\basex{0.0\phantom{0}}} & 55.9 \phantom{\basex{0.0\phantom{0}}} & 56.4 \phantom{\basex{0.0\phantom{0}}}\\
    AimCLR~\cite{guo2022contrastive} & Joint & 74.3 \basexx{6.0\phantom{0}}& 79.7 \basexx{3.3\phantom{0}}& 77.0 \basexx{4.6\phantom{0}}& 63.4 \basexx{6.6\phantom{0}}& 63.4 \basexx{7.5\phantom{0}}& 63.4 \basexx{7.0\phantom{0}}\\
    ActCLR~\cite{lin2023actionlet} & Joint & 80.9 \basexx{12.6} & 86.7 \basexx{10.3} & 83.8 \basexx{11.4} & 69.0 \basexx{12.2} & 70.5 \basexx{14.6} & 69.8 \basexx{13.4}\\
    \textbf{TranCLR} & Joint & \textbf{83.8} \up{15.5}& \textbf{87.9} \up{11.5}& \textbf{85.9} \up{13.5}& \textbf{74.3} \up{17.5}& \textbf{74.7} \up{18.8}& \textbf{74.5} \up{18.1}\\

    \midrule
    
    SkeletonCLR~\cite{li20213d} & Motion & 53.3 \phantom{\basex{0.0\phantom{0}}} & 50.8 \phantom{\basex{0.0\phantom{0}}} & 52.1 \phantom{\basex{0.0\phantom{0}}} & 39.6 \phantom{\basex{0.0\phantom{0}}} & 40.2 \phantom{\basex{0.0\phantom{0}}} & 39.9 \phantom{\basex{0.0\phantom{0}}}\\
    AimCLR~\cite{guo2022contrastive} & Motion & 66.8 \basexx{13.5} & 70.6 \basexx{19.8} & 68.7 \basexx{16.6} & 57.3 \basexx{17.7} & 54.4 \basexx{14.2} & 55.9 \basexx{16.0}\\
    ActCLR~\cite{lin2023actionlet} & Motion & 78.6 \basexx{25.3} & 84.4 \basexx{33.6} & 81.5 \basexx{29.4} & 68.3 \basexx{28.7} & 67.8 \basexx{27.6} & 68.1 \basexx{28.2}\\
    \textbf{TranCLR} & Motion & \textbf{81.2} \up{27.9} & \textbf{84.8} \up{34.0} & \textbf{83.0} \up{30.9} & \textbf{71.4} \up{31.8} & \textbf{70.3} \up{30.1} & \textbf{70.9} {\up{31.0}}\\

    \midrule
    
    SkeletonCLR~\cite{li20213d} & Bone & 69.4 \phantom{\basex{0.0\phantom{0}}} & 67.4 \phantom{\basex{0.0\phantom{0}}} & 68.4 \phantom{\basex{0.0\phantom{0}}} & 48.4 \phantom{\basex{0.0\phantom{0}}} & 52.0 \phantom{\basex{0.0\phantom{0}}} & 50.2 \phantom{\basex{0.0\phantom{0}}}\\
    AimCLR~\cite{guo2022contrastive} & Bone & 73.2 \basexx{3.8\phantom{0}} & 77.0 \basexx{9.6\phantom{0}} & 75.1 \basexx{6.7\phantom{0}} & 62.9 \basexx{14.5} & 63.4 \basexx{11.4} & 63.2 \basexx{13.0}\\
    ActCLR~\cite{lin2023actionlet} & Bone & 80.1 \basexx{10.7} & 85.0 \basexx{17.6} & 82.6 \basexx{14.2} & 67.8 \basexx{19.4} & 68.2 \basexx{16.2} & 68.0 \basexx{17.8}\\
    \textbf{TranCLR} & Bone & \textbf{83.3} \up{13.9} & \textbf{87.1} \up{19.7} & \textbf{85.2} \up{16.8} & \textbf{73.9} \up{25.5} & \textbf{73.3} \up{21.3} & \textbf{73.6} \up{23.4}\\

    \midrule

    3s-SkeletonCLR~\cite{li20213d} & \makebox[3cm][c]{Joint+Motion+Bone} & 75.0 \phantom{\basex{0.0\phantom{0}}} & 79.8 \phantom{\basex{0.0\phantom{0}}} & 77.4 \phantom{\basex{0.0\phantom{0}}} & 60.7 \phantom{\basex{0.0\phantom{0}}} & 62.6 \phantom{\basex{0.0\phantom{0}}} & 61.7 \phantom{\basex{0.0\phantom{0}}}\\
    3s-AimCLR~\cite{guo2022contrastive} & \makebox[3cm][c]{Joint+Motion+Bone} &  78.9 \basexx{3.9\phantom{0}} & 83.8 \basexx{4.0\phantom{0}} & 81.4 \basexx{4.0\phantom{0}} & 68.2 \basexx{7.5\phantom{0}} & 68.8 \basexx{6.2\phantom{0}} & 68.5 \basexx{6.8\phantom{0}}\\
    3s-ActCLR~\cite{lin2023actionlet} & \makebox[3cm][c]{Joint+Motion+Bone} & 84.3 \basexx{9.3\phantom{0}} & 88.8 \basexx{9.0\phantom{0}} & 86.6 \basexx{9.2\phantom{0}} & 74.3 \basexx{13.6} & 75.7 \basexx{13.1} & 75.0 \basexx{13.3}\\
    \textbf{3s-TranCLR} & \makebox[3cm][c]{Joint+Motion+Bone} & \textbf{86.3} \up{11.3} & \textbf{90.7} \up{10.9} & \textbf{88.5} \up{11.1} & \textbf{78.8} \up{18.1} & \textbf{79.0} \up{16.4} & \textbf{78.9} \up{17.2}\\
    \bottomrule
\end{tabular}
\label{tab1}
\vspace{-10pt}
\end{table*}

\section{Experiments}

\subsection{Experimental Settings}

To ensure a fair comparison with ActCLR~\cite{lin2023actionlet}, our encoder $f(\cdot)$ is based on ST-GCN with 16 hidden channels, which is a quarter of the size of the original ST-GCN model. All experiments are conducted on an NVIDIA A100 GPU. The batch size for both pretraining and downstream tasks is set to 128 by default. 


\mypara{Self-Supervised Pretext Training.} The projection head $g(\cdot)$ is a 2-layer MLP that projects features from 256 dimensions to 128 dimensions. For the soft alignment, the temperatures are set to $\tau_q=0.1$ and $\tau_k=0.05$, with $K=8192$.  We utilize a memory bank of size 65536. For optimization, we use SGD with momentum 0.9 and weight decay 0.0001. The model is trained for 300 epochs with an initial learning rate of 0.1, which decreases to 0.01 at epoch 250.

\subsection{Evaluation Protocols}

To comprehensively evaluate our self-supervised encoder, we assess the learned representations in terms of discriminability, generalization, semantic structure, and reliability.

\mypara{Linear Evaluation.} To evaluate the discriminative power of the learned representations, we freeze the pretrained encoder, append a linear classifier (a \emph{fully-connected} layer followed by \emph{softmax}), and train it in a supervised manner. 

\mypara{Transfer Learning.} To assess the generalization ability of our learned representations, we first pretrain the encoder on the source dataset, and then evaluate it on a target dataset using the linear evaluation protocol above. 

\mypara{Skeleton-Based Action Retrieval.} This protocol evaluates the semantic structure of the learned embedding space. For a query skeleton sequence, we extract its feature with the frozen pre-trained encoder, retrieve the nearest neighbor from the training set based on cosine similarity, and assign that neighbor's label to the query.

\mypara{Calibration Analysis.} Model calibration measures the alignment between a model's predicted confidence scores and its actual accuracy. This can be quantified by the Calibration Error~\cite{naeini2015obtaining}, defined as:
\begin{equation*}\mathbb{E}_{(\boldsymbol{x}_i,y_i)\sim \mathcal{D}} \left[ \left\vert P(\hat{y}_i = y_i \mid \hat{p}) - \hat{p} \right\vert \right].
\end{equation*}
To our knowledge, calibration remains underexplored in self-supervised human action recognition. We introduce it to evaluate the reliability of learned representations. Following the linear evaluation protocol, we compute the Expected Calibration Error (ECE)~\cite{guo2017on} and Adaptive ECE (AECE)~\cite{ding2020revisiting} of the downstream classifier, hypothesizing that a better feature space yields a well-calibrated model.

\subsection{Comparison with State-of-the-Art Methods}

\mypara{Linear Evaluation Results.} 
As shown in \cref{tab1}, TranCLR consistently establishes new state-of-the-art performance across all settings on the NTU datasets. In single-stream comparisons, our model significantly outperforms prior methods. For instance, on the challenging NTU-120 X-Sub, our joint-based model achieves 74.3\% accuracy, surpassing the strong contemporary baseline ActCLR~\cite{lin2023actionlet} by a remarkable 5.3\%. This advantage is amplified in the multi-stream setting, where 3s-TranCLR achieves an unprecedented 88.5\% average accuracy on NTU-60 and 78.9\% on NTU-120, demonstrating the powerful synergy of our calibrated representations across different modalities. Furthermore, as presented in \cref{tab2}, our method's superiority extends to the PKU-MMD dataset, where it outperforms a diverse array of competitors spanning different paradigms. It is particularly noteworthy that on the challenging Part II, 3s-TranCLR sets a new state-of-the-art at 59.9\%, underscoring its robustness. These compelling results validate that by explicitly modeling action transitions via ATAC and enforcing multi-level geometric calibration, TranCLR learns exceptionally high-quality representations that excel in downstream classification tasks. 


\begin{table}[tb]
  \small
  \centering
  \caption{Comparison of linear evaluation results for self-supervised learning approaches on the PKU-MMD dataset.}
  \begin{tabular}{llcc}
    \toprule
   \multirow{2}{*}{\textbf{Method}} & \multirow{2}{*}{\textbf{Publication}} & \multicolumn{2}{c}{\textbf{PKU-MMD}}\\
   \cmidrule(lr){3-4}
    & & Part I  & Part II\\
    \midrule

    \rowcolor{gray!20} \multicolumn{4}{l}{\textbf{\emph{Feature Decorrelation}}}\\
    HYSP~\cite{franco2023hyperbolic} & ICLR$'$23 & 88.8 & -\\
    3s-UmURL~\cite{sun2023unified} & MM$'$23 & - & 54.3\\
    3s-USDRL~\cite{weng2025usdrl} & AAAI$'$25 & - & 59.7\\
    Heter-Skeleton~\cite{wang2025heterogeneous} & CVPR$'$25 & - & 58.2\\

    \rowcolor{gray!20}\multicolumn{4}{l}{\textbf{\emph{Masked Sequence Modeling}}}\\
    MAMP~\cite{mao2023mamp} & ICCV$'$23 & 92.2 & 53.8 \\
    SkeletonMAE~\cite{wu2023skeletonmae} & ICMEW$'$23 & 82.8 & 36.1 \\
    3s-PCM$^3$~\cite{zhang2023prompted} & MM$'$23 & 86.3 & 58.2 \\
    MacDiff~\cite{wu2024macdiff} & ECCV$'$24 & \textbf{92.8} & -\\
    S-JEPA~\cite{abdelfattah2024sjepa} & ECCV$'$24 & 92.2 & 53.5\\
    GFP~\cite{sun2025efficient} & ICCV$'$25 & - & 56.2\\
    NAT-CL~\cite{tao2025rethinking} & AAAI$'$25 & - & 55.3\\
    STARS~\cite{mehraban2025stars} & WACV$'$26 & 91.2 & 52.7\\

    \rowcolor{gray!20}\multicolumn{4}{l}{\textbf{\emph{Contrastive Learning}}}\\
    3s-CrosSCLR~\cite{li20213d} & CVPR$'$21 & 84.9 & 21.2\\
    3s-AimCLR~\cite{guo2022contrastive} & AAAI$'$22 & 87.4 &  39.5 \\
    3s-ActCLR~\cite{lin2023actionlet} & CVPR$'$23 & 91.9 &  55.8 \\
    HiCo~\cite{hico2023} & AAAI$'$23 & 89.4 & 54.7\\
    HaLP~\cite{shah2023halp} & CVPR$'$23 & - & 43.5\\
    3s-SkeAttnCLR~\cite{Hua2023SkeAttnCLR} & IJCAI$'$23 & 89.5 & 55.5\\
    3s-PSTL~\cite{zhou2023self} & AAAI$'$23 & 89.2 & 52.3\\
    I$^2$MD~\cite{mao2023i2md} & IJCV$'$24 & - & 49.0\\
    A$^2$MC~\cite{xu2024attack} & TPAMI$'$24 & 92.5 & -\\
    3s-Skeleton-logoCLR~\cite{hu2024global} & TCSVT$'$24 & 92.2 & 57.7\\
    PASTD~\cite{qian2025pastd} & ICASSP$'$25 & 92.0 & 54.6\\

    \midrule

    \textbf{3s-TranCLR} & This work & \textbf{92.8} & \textbf{59.9} \\

    \bottomrule
\end{tabular}
  \label{tab2}
  \vspace{-14pt}
\end{table}

\mypara{Transfer Learning Results.} As presented in \cref{tab3}, 3s-TranCLR exhibits remarkable transferability. When pretrained on NTU-60, it achieves a state-of-the-art accuracy of 65.6\%, significantly outperforming recent strong competitors by 1.3\% over Heter-Skeleton~\cite{wang2025heterogeneous} and 3.5\% over the reconstruction-enhanced 3s-ActCLR+~\cite{lin2025self}. We attribute this superior generalization to our MGMC mechanism. By constructing a smoother and more continuous action manifold, TranCLR avoids overfitting to the source domain's specific data distribution. This encourages the model to learn more fundamental and transferable motion patterns, demonstrating its practical value for leveraging large-scale unlabeled data in real-world scenarios.

\mypara{Skeleton-Based Action Retrieval Results.} As reported in \cref{tab4}, TranCLR achieves surprising performance, setting new state-of-the-art results on the NTU-60 X-Sub (74.6\%) and NTU-120 X-Sub (59.1\%). These results confirm that our method produces a highly discriminative feature space where actions are well-separated. While TranCLR does not lead on every metric, this is an expected outcome of our core design philosophy. By intentionally softening class boundaries via MGMC mechanisms to construct a continuous action manifold, we sacrifice peak retrieval precision that depends on rigid separation to gain better generalization and topological coherence. Crucially, despite this inherent trade-off, our model achieves state-of-the-art results on multiple benchmarks, powerfully validating the quality of the learned representations and our approach's unique balance of discriminability with semantic structure.

{\setlength{\tabcolsep}{18pt}
\begin{table}[tb]
  \small
  \centering
  \caption{Comparison of transfer learning performance on PKU-MMD Part II, with models pretrained on NTU X-Sub benchmark.}
  \begin{tabular}{@{\hspace{7pt}}lcc}
    \toprule
   \multirow{2}{*}{\textbf{Method}}& \multicolumn{2}{c}{\textbf{Transfer to PKU-MMD II}}\\
   \cmidrule{2-3}
    &   \textbf{NTU-60} & \textbf{NTU-120}\\
    \midrule
    HiCo~\cite{hico2023}  & 56.3  & 55.4 \\
    CMD~\cite{Mao_2022_CMD}  & 56.0  & 57.0 \\
    3s-AimCLR~\cite{guo2022contrastive}  & 51.6 & 52.7 \\
    UmURL~\cite{sun2023unified}  & 59.7  & 58.5 \\
    USDRL~\cite{weng2025usdrl}  & 57.2 & 58.3 \\
    SkeletonMAE~\cite{wu2023skeletonmae} & 58.4 & 61.0 \\
    3s-ActCLR~\cite{lin2023actionlet}  & 55.9 & 59.6\\
    3s-ActCLR+~\cite{lin2025self}  & 62.1 & -\\
    Heter-Skeleton~\cite{wang2025heterogeneous}  & 64.3  & 63.1 \\
    \midrule
\textbf{3s-TranCLR} & \textbf{65.6}  & \textbf{64.9}\\
    \bottomrule
\end{tabular}
  \label{tab3}
\end{table}
}

\begin{table}[tb]
	\centering
	\caption{Comparison of skeleton-based action retrieval performance with unsupervised learning approaches on the NTU dataset.}
  \begin{tabular}{lc@{\hspace{6pt}}cc@{\hspace{6pt}}c}
			\toprule
			\multirow{2}{*}{\textbf{Method}} & \multicolumn{2}{c}{\textbf{NTU-60}} & \multicolumn{2}{c}{\textbf{NTU-120}} \\
			\cmidrule(lr){2-3} \cmidrule(lr){4-5}   
      & X-Sub & X-View & X-Sub & X-Set \\
			\midrule
			ISC~\cite{fmthoker_acmmm}   & 62.5  & 82.6  & 50.6  & 52.3 \\
			HaLP~\cite{shah2023halp}  & 65.8  & 83.6  & 55.8  & 59.0 \\
			CMD~\cite{Mao_2022_CMD}   & \underline{70.6}  & \underline{85.4}  & \underline{58.3}  & \underline{60.9} \\
			HiCo~\cite{hico2023}  & 68.3  & 84.8  & 56.6  & 59.1 \\
			ActCLR~\cite{lin2023actionlet} & 66.6 & 78.0 & 49.0 & 52.6 \\
			Heter-Skeleton~\cite{wang2025heterogeneous} &  66.3 & \textbf{87.1} & 55.7 & 59.8 \\
			\midrule
      \textbf{TranCLR} & \textbf{74.6} & 83.9 & \textbf{59.1} & \textbf{62.5} \\
			\bottomrule
	\end{tabular}
	\label{tab4}
    \vspace{-12pt}
\end{table}

\mypara{Calibration Analysis Results.} The results in \cref{tab5} are striking: TranCLR dramatically improves model calibration. On the NTU-120 X-Set, our method slashes the ECE from 5.63\% (ActCLR~\cite{lin2023actionlet}) to a near-perfect 0.65\%, achieving roughly an 88\% relative reduction. We attribute this substantial gain directly to our MGMC mechanism. By enforcing intra-, inter-, and cross-sample consistency, MGMC constructs a highly regularized manifold that prevents the downstream classifier from becoming overconfident. This enhanced reliability is vital for real-world applications requiring trustworthy uncertainty estimation, highlighting the practical value of TranCLR. 


{\setlength{\tabcolsep}{7.5pt}
\begin{table}[tb]
  \small
  \centering
  \caption{Comparison of model calibration performance under linear evaluation on the NTU dataset.}
  \begin{tabular}{l@{\hskip 6pt}lc@{\hskip 5pt}cc@{\hskip 5pt}c}
    \toprule
    \multirow{2}{*}{\textbf{Metric}} & \multirow{2}{*}{\textbf{Method}} & \multicolumn{2}{c}{\textbf{NTU-60}} & \multicolumn{2}{c}{\textbf{NTU-120}}  \\
    \cmidrule(lr){3-4} \cmidrule(lr){5-6}
    & & X-Sub & X-View & X-Sub & X-Set\\
    \midrule
    \multirow{3}{*}{\textbf{ECE$\downarrow$  }} & AimCLR~\cite{guo2022contrastive} & 7.11 & 4.85 & 7.64 & 7.71 \\
    &  ActCLR~\cite{lin2023actionlet} & 5.25 & 2.51 & 5.71 & 5.63 \\
    &  \textbf{TranCLR} & \textbf{0.98} & \textbf{1.27} & \textbf{0.78} & \textbf{0.65} \\
    \midrule
    \multirow{3}{*}{\textbf{AECE$\downarrow$ }} & AimCLR~\cite{guo2022contrastive} & 7.10 & 4.85 & 7.64 & 7.71 \\
    &  ActCLR~\cite{lin2023actionlet} & 5.25 & 2.47 & 5.71 & 5.63 \\
    &  \textbf{TranCLR} & \textbf{0.87} & \textbf{1.27} & \textbf{0.67} & \textbf{0.64} \\
    \bottomrule
\end{tabular}
  \label{tab5}
\end{table}
}

{\setlength{\tabcolsep}{10pt}
\begin{table}[tb]
  \small
  \centering
  \caption{Ablation study on ATAC components.}
  \begin{tabular}{cc|ccc}
    \toprule
    \multirow{2}{*}{\textbf{w/ Global}} & \multirow{2}{*}{\textbf{w/ Local}} &  \multicolumn{3}{c}{\textbf{NTU-60}} \\
    & & X-Sub & X-View & Avg.\\
    \midrule
    \ding{55} & \ding{55} & 74.9 & 79.9 & 77.4 \\
    \ding{51} & & 81.5 & 86.0 & 83.8\\
    & \ding{51} & 83.3 & 86.8 & 85.0\\
    \ding{51} & \ding{51} & \textbf{83.8} & \textbf{87.9} & \textbf{85.9}\\
    \bottomrule
  \end{tabular}
  \label{tab6}
\end{table}
}

\begin{table}[tb]
\centering
\small
\caption{Ablation study on MGMC calibration levels.}
\begin{tabular}{ccc|ccc}
\toprule
\multirow{2}{*}{\textbf{w/ $\mathcal{L}_{\text{intra}}$}} & \multirow{2}{*}{\textbf{w/ $\mathcal{L}_{\text{inter}}$}} & \multirow{2}{*}{\textbf{w/ $\mathcal{L}_{\text{cross}}$}} & \multicolumn{3}{c}{\textbf{NTU-60}} \\
& & & X-Sub & X-View & Avg. \\
\midrule
\ding{55} & \ding{55} & \ding{55} & 74.9 & 79.9 & 77.4 \\
\ding{51} &  &  & 81.2 & 84.8 & 83.0 \\
 & \ding{51} &  & 77.3 & 80.1 & 78.7 \\
\ding{51} & \ding{51} &  & 83.0 & 86.5 & 84.8 \\
\ding{51} & \ding{51} & \ding{51} & \textbf{83.8} & \textbf{87.9} & \textbf{85.9} \\
\bottomrule
\end{tabular}
\label{tab7}
\vspace{-12pt}
\end{table}

\subsection{Ablation Study}

\mypara{Effectiveness of ATAC.} We evaluate the components of our ATAC mechanism on the NTU-60 dataset, with results presented in \cref{tab6}. Using only global interpolation yields the lowest average accuracy of 83.8\%, indicating that while it promotes manifold smoothness, it blurs the fine-grained discriminative kinematics essential for capturing detailed motion patterns. Conversely, omitting global interpolation results in a significantly stronger average performance of 85.0\%, showcasing the power of our proposed local substitution mechanism even in isolation. However, this result still falls short of the full model, suggesting that the local-only view struggles to enforce a globally coherent structure.
The complete ATAC framework, which dynamically integrates both strategies, achieves the highest average accuracy of 85.9\%. This confirms that the synergy between enforcing global smoothness and preserving local discriminability is indispensable for learning a holistically well-structured action manifold.

\begin{table}[tb]
	\centering
	\small
	\caption{Ablation study on Local Substitution parameters.}
	\begin{tabular}{ccccc|c@{\hskip 6.5pt}c@{\hskip 6.5pt}c}
		\toprule
		\multirow{2}{*}{$\mathcal{S}_{\text{min}}$} & \multirow{2}{*}{$\mathcal{S}_{\text{max}}$} & \multirow{2}{*}{$\mathcal{T}_{\text{min}}$} & \multirow{2}{*}{$\mathcal{T}_{\text{max}}$} & \multirow{2}{*}{$[\kappa_l, \kappa_r]$} & \multicolumn{3}{c}{\textbf{NTU-60}} \\
		& & & & & X-Sub & X-View & Avg.\\
		\midrule
        - & - & - & - & - & 74.9 & 79.9 & 77.4 \\
        1 & 2 & 8 & 16 & $[1,1]$ & 82.8 & 87.0 & 84.9 \\
        2 & 3 & 8 & 16 & $[1,1]$ & 83.1 & 87.2 & \textbf{85.2} \\
        3 & 4 & 8 & 16 & $[1,1]$ & 82.9 & 87.1 & 85.0 \\
\midrule
        2 & 3 & 8 & 16 & $[1,1]$ & 83.1 & 87.2 & 85.2 \\
        2 & 3 & 16 & 24 & $[1,1]$ & 83.3 & 87.7 & \textbf{85.5} \\
        2 & 3 & 24 & 32 & $[1,1]$ & 83.0 & 87.4 & 85.2 \\
		\midrule
		2 & 3 & 16 & 24 & $[\phantom{.}1\phantom{0},1]$ & 83.3 & 87.7 & 85.5\\
		2 & 3 & 16 & 24 & $[0.5,1]$ & 83.7 & 87.7 & 85.7\\
		2 & 3 & 16 & 24 & $[\phantom{.}1\phantom{0},2]$ & 83.6 & 87.5 & 85.5\\
        2 & 3 & 16 & 24 & $[0.5,2]$ & 83.8 & 87.9 & \textbf{85.9}\\
		\bottomrule
	\end{tabular}
	\label{tab8}
    \vspace{-12pt}
\end{table}

\mypara{Effectiveness of MGMC.} We dissect our MGMC framework, with results detailed in \cref{tab7}. Using only intra-sample calibration establishes a strong foundation, achieving an average accuracy of 83.0\%. Inter-sample calibration alone also brings improvements over the baseline, reaching an average accuracy of 78.7\%. However, it is less effective than intra-sample calibration, indicating that solely bridging different action classes is limited without a stable intra-sample constraint. When inter-sample calibration is built on this foundation, it provides a substantial boost to 84.8\%, proving its synergistic value. Finally, adding cross-anchor relational consistency as a global regularizer refines the manifold to achieve our best average accuracy of 85.9\%. This progression shows that the combined effects of intra-sample continuity, inter-sample bridging, and cross-anchor consistency are essential for constructing a coherent and topologically consistent feature space.

\mypara{Parameter Analysis for Local Substitution.} We analyze the hyperparameters for Local Spatio-temporal Substitution in \cref{tab8}. Optimal performance is achieved by substituting 2$\sim$3 body parts, which is consistent with biomechanics, as actions such as \emph{eating} or \emph{walking} typically involve the coordination of 2$\sim$3 limbs. This works best with a temporal window of $[16, 24]$ frames, which is sufficient to capture a complete sub-action without introducing irrelevant context. Flexible scaling with $\kappa \in [0.5, 2]$ further boosts robustness to natural speed variations, yielding a peak average accuracy of 85.9\%. Based on this empirical analysis, we adopt this configuration as our default, as it is not arbitrary but grounded in the physical principles of human motion. 


\vspace{-1mm}
\section{Conclusion}
In this work, we present TranCLR, a transitional anchor-based contrastive learning framework for skeleton-based action understanding. By introducing action transitional anchors as manifold regularizers, TranCLR enriches the latent topology of human actions. We further design a multi-level geometric manifold calibration mechanism to preserve global topological coherence and promote a smooth, discriminative representation space. Extensive experiments on multiple challenging datasets show that TranCLR consistently outperforms state-of-the-art methods in accuracy, generalization, and confidence calibration.

\section*{Acknowledgments}
This work was supported by the Guangdong Basic and Applied Basic Research Foundation (2025A1515011546) and the Shenzhen Science and Technology Program (ZDCY20250901113000001). It was also supported in part by the Joint Funds of the National Natural Science Foundation of China under Grant U24A20248.
{
    \small
    \bibliographystyle{ieeenat_fullname}
    \bibliography{main}
}


\end{document}